\documentclass[runningheads]{llncs}
\usepackage{graphicx}
\usepackage{subcaption}
\usepackage{caption}
\usepackage{xspace} 
 \usepackage[T1]{fontenc}
\graphicspath{{Results/}}
\usepackage{amssymb}
\newcommand{\mpara}[1]{\medskip\noindent{\bf #1}}
\newcommand{\deepwalk}{\textsc{DeepWalk}\xspace}
\newcommand{\app}{\textsc{APP}\xspace}
\newcommand{\hope}{\textsc{HOPE}\xspace}
\newcommand{\graphsage}{\textsc{GraphSage}\xspace}
\newcommand{\nerd}{\textsc{NERD}\xspace}
\newcommand{\glae}{\textsc{Glae}\xspace}
\newcommand{\wordtovec}{\textsc{word2vec}\xspace}

\begin{document}
\title{Graph-based State Representation for Deep Reinforcement Learning}


\author{Vikram Waradpande \and
Daniel Kudenko\and Megha Khosla }
\authorrunning{V. Warapande et al.}
%
\institute{L3S Research Center, Leibniz University, Hannover\\
\email{\{waradpande,khosla,kudenko\}@l3s.de}}
\maketitle              
\begin{abstract}

Deep RL approaches build much of their success on the ability of the deep neural network to generate useful internal representations. Nevertheless, they suffer from a high sample-complexity and starting with a good input representation can have a significant impact on the performance. In this paper, we exploit the fact that the underlying Markov decision process (MDP) represents a graph, which enables us to incorporate the topological information for effective state representation learning.

Motivated by the recent success of node representations for several graph analytical tasks we specifically investigate the capability of node representation learning methods to effectively encode the topology of the underlying MDP in Deep RL. To this end we perform a comparative analysis of several models chosen from 4 different classes of representation learning algorithms for policy learning in grid-world navigation tasks, which are representative of a large class of RL problems. We find that all embedding methods outperform the commonly used matrix representation of grid-world environments in all of the studied cases. Moreoever, graph convolution based methods are outperformed by simpler random walk based methods and graph linear autoencoders. 


\end{abstract}
\section{Introduction}

A good problem representation has been known to be crucial for the performance of AI algorithms. This is not different in the case of reinforcement learning (RL), where representation learning has been a focus of investigation. The core idea is to map the high dimensional state space to low dimensional latent representations which are more informative thus facilitating the learning of an optimal policy. Previous works, for example, \cite{jonschkowski2015learning,munk2016learning} have focused on learning these representations from the incoming high dimensional signals or observations corresponding to a state but have ignored the fact that the underlying stochastic decision process induces a topological structure over the states which can provide additional useful features. In this work, we focus on extracting informative low dimensional state features based on the topological structure of the underlying Markov decision processes (MDPs) for deep reinforcement learning. 




Deep reinforcement learning combines neural networks with a reinforcement learning architecture. In particular, rather than using a lookup table 
a neural network is used to approximate a value function (and thus a policy) without the need to store, index and update all possible states and their values in look-up tables. While Deep RL approaches build much of their success on the ability of the deep neural network to generate useful internal representations, they nevertheless suffer from a high sample complexity. One way to overcome this problem, is to start with a better input representation that can improve the learning performance significantly. In this paper, we exploit the fact that the underlying Markov decision process (MDP) represents a graph and investigate the suitability of graph representation learning approaches to learn effective encodings of the state. These encodings are then used to enrich the sate representation for Deep RL and improve the speed of learning.


Graph Representation learning (GRL) approaches~\cite{cai2018comprehensive,perozzi2014deepwalk,hamilton2017inductive,ou2016asymmetric} aim to embed nodes in a low dimensional space such that the topological structure of the graph is preserved. Though these methods have gained popularity and showed state of the art improvements in several graph analytical tasks like node classification and link prediction, their suitability or generalizability to different domains has escaped attention so far. Our work identifies Deep RL for discrete MDPs as a promising application for utilizing and evaluating graph representations. More specifically, we evaluate several unsupervised representation learning methods on their ability to learn effective state representations encoding the topological structure of MDP. With a large number of unsupervised GRL approaches proposed to date we systematically choose models from 4 GRL classes based on several criteria as elaborated in Section~\ref{sec:rational}. We then investigate \textit{which approaches are best suited to be used as state representation learning methods for MDPs in Deep RL.}

While the node embeddings are computed from a given MDP, it is important to note that in this work we do not assume that the entire MDP is known in advance. Instead, we first generate an estimate of MDP by collecting random samples and use the resulting graph to generate node representations in an unsupervised manner. We do assume that the environment is discrete, since otherwise it is not possible to represent the MDP as a finite graph. In the case of continuous states, these could be discretised beforehand, either manually or via an automated method analogous to tile coding. 

\mpara{Our Contributions.} We evaluate GRL approaches for effectively encoding MDPs under the assumption that the entire MDP is not known in advance. To the best of our knowledge, our work is the first to propose the use of pretrained state representations entirely on the topological structure of the MDP and show that dense low dimensional input state representations enhance the performance of Deep Q-Networks (DQN) for navigational tasks. 
To summarize our main contributions are as follows.
\begin{itemize}
    \item We propose and evaluate a wide range of graph based representation learning approaches to generate state features based on topological structure of MDPs, leading to improved learning performance in DeepRL.
    \item We show that RL is a promising application for evaluating and enhancing graph representation learning approaches.
\end{itemize}
\mpara{Key Findings.} Our key findings and conclusions are
\begin{itemize}
    \item Pre-trained unsupervised low-dimensional state representations when used as input to DQN shows state of the art improvements over the raw high dimensional state input for grid-world environments.
    \item  For undirected MDPs, quite surprisingly, the first neural network based representation learning method \deepwalk outperforms almost all other methods including the more popular graph convolution based methods.
    \item By varying the number of samples used to generate an approximate MDP  we show that the best performing embedding approaches show comparable performance even with smaller number of samples.
    \item For directed MDPs, preserving edge directionality while learning state representations does not appear to be crucial. For instance \deepwalk, when used to train representations while ignoring the edge  directionality, showed comparable performance to \app and \nerd which generated representations in directionality preserving manner.
\end{itemize}
\section{Preliminaries and Related Work}
\subsection{Markov Decision Processes in RL }
Markov  decision  processes (MDP)  are  discrete  time  stochastic control processes which are used to formalize reinforcement learning problems and model RL environments. MDPs are represented by a 4-tuple $(S,A,P,R)$, where $S$ is a set of discrete states an agent can be in, $A$ is the set of all possible actions that the agent can take, $P$ denotes a probability density function with $P(s’|s,a)$ being the transition probability of moving from state $s$ to state $s'$ after taking action $a$, and $R(s,a)$ is the immediate reward that the agent receives when it takes action $a$ in state $s$. The objective of reinforcement learning is then to determine the optimal mapping of a given state to action, $\pi(s)$ (the {\it policy}) such that the chosen action results in maximizing the expected sum of rewards received in the future.






\subsection{DQN}

A widely used modern RL algorithm is Deep Q-Networks (DQN) \cite{DQN}. DQN is the "deep" expansion of Q-Learning \cite{RLIntro} and uses essentially the same update rules and operating principles as Q-Learning but adapted to use a neural network as its value function representation.

Specifically, the DQN algorithm computes the Q value function (represented as a deep neural network). This value function maps a state $s$ and action $a$ into an estimate of the expected cumulative reward for executing action $a$ in state $s$ and following the optimal policy from then on. As the agent interacts with its environment, the agent accumulates experience in the form of $(s, a, r, s')$ tuples, which are used to update the neural network that computes the Q function. For each experience tuple, the values $Q(s,a)$ and $r + \max_{a'}{Q(s',a')}$ are calculated using the neural network. The difference between these two values is used in the loss function to update the network. This "bootstrapping" method enables the agent to learn strong estimates for the expected return of each state-action pair. 

The main strength of DQN is that the deep neural network can generate useful internal representations of environment states when provided with very simple state representations. However, DQN suffers from a very high sample complexity. The complexity can be reduced by generating a more effective state representation before feeding it into the neural network, and this forms the basis of the approach proposed in this paper. 


\subsection{Graph Representations} Graph representation learning (GRL) aims to learn low-dimension latent representations of nodes to be used for downstream tasks such as link prediction, node classification etc. GRL methods include \textit{random walk} based methods, \textit{matrix factorization} based and \textit{graph neural networks} (GNNs). Random walk based~\cite{perozzi2014deepwalk,NERD,zhou2017scalable} methods optimize the node embeddings so that nodes have similar embeddings if they tend to co-occur on random walks over the graph. Matrix factorization based~\cite{ou2016asymmetric} methods rely on low rank decomposition of a target matrix such as the $k$-step transition probability matrix, modularity matrix etc. to obtain node encodings. GNNs are deep learning models designed to extract features from the graph structure as well as the input node attributes and can be further categorized into recurrent graph neural networks~\cite{scarselli2008graph}, convolutional graph neural networks~\cite{kipf2017semi,niepert2016learning} and graph autoencoders~\cite{kipf2016variational,salha2019keep}. 
In  spite  of  their  success,  there are limited studies on in-depth comparative analysis~\cite{khosla2019comparative,errica2020a} of these methods over a wide range of datasets and tasks. Moreover, none of these works focus on generalizability of embedding approaches to encode graph structure beyond using the encodings in node/graph classification or link prediction tasks. 

\subsection{State Representations in Reinforcement Learning} In reinforcement learning, it is a common practice to map the state(-action) space to a low dimensional latent space where the main goal of such a transformation is to represent the input data in a more informative form that facilitates and improves subsequent steps. The authors in \cite{osentoski2007learning} and \cite{mahadevan2007proto} proposed the use of the  Laplacian basis functions as state encodings. Basis
functions are derived by finding the “smoothest” eigen vectors (that correspond to the smallest eigen values) of the graph Laplacian and is argued that such smooth eigenvectors also reflect the smoothness of value functions over nearby nodes. On the one hand eigenvalue decomposition is computationally expensive and on the other hand the smoothness assumption for value functions over the states might not be always valid as also observed in \cite{madjiheurem2019representation} . We note that \cite{madjiheurem2019representation} also emphasize the use of node embeddings as basis functions to be used in a generalized version of representation policy iteration (RPI).
The authors in \cite{jonschkowski2015learning} use priors about the structure of robotic interactions with the physical world to learn state representations. Our work can be seen as complementary in nature as we propose a more general approach of encoding such an observed structure into state-action representations. Other works~\cite{de2018integrating,munk2016learning} which perform state representation learning and reinforcement learning simultaneously have also ignored the fact that the underlying decision process represents a graph and is therefore crucial to incorporate the topological information for effective state representation learning. Recently \cite{jiang2018graph} proposed a graph convolutional network based approach to multi-agent reinforcement learning.
But it is not known a priori if the chosen graph learning model is the best for the considered use cases. In this work we are evaluating the capability of graph learning methods on effectively learning from the MDP structure, focusing on discrete navigation problems (i.e. mazes) which are representative for a large class of MDPs. The learnt representations can be further refined and combined with more input features like sensor observations corresponding to states for more specific cases and corresponding architectures and objective functions for example in \cite{de2018integrating,munk2016learning} can be employed.

\section{Comparative Analysis}
Unsupervised GRL approaches aim to learn low dimensional representations for each node while preserving certain topological characteristics of the underlying graph. Informally, each of the methods tend to encode \textit{similar} nodes closer in the embedding space and the definition of similarity varies from methods to method. Consequently we might expect that there cannot be a single winner method which can encode all the types of graph structure well and the choice of a particular method would depend on certain structural characteristics of the underlying graph which might be application specific. Therefore we choose  a wide range of representative methods from several classes of GRL methods and argue about their suitability to be used in Reinforcement learning applications.


\subsection{Compared Models}
 Let $G=(V,E)$ denotes the graph with $|V|$ nodes and $|E|$ edges corresponding to the MDP estimated from the obtained random samples. We train low dimensional vector representation or embedding, $\phi(v)$ for each node $v \in V$ using the following GRL methods.
 
\mpara{\deepwalk\cite{perozzi2014deepwalk}.} Inspired by techniques of language modelling and unsupervised feature learning from word-sequences \deepwalk generates node-sequences from graphs using short random walks and trains the Skip-Gram model using hierarchical softmax akin to \wordtovec-based training procedure. The training set is prepared by sampling vertex-context pairs over a sliding window in a given random walk. In particular, it attempts to find node embeddings such that the likelihood of observing a vertex $v_i \in V$ given its context ( i.e. other neighboring vertices within a specified window of the random walk) is maximized. Let $\Phi$ denotes the latent representation matrix, $\{v_{i-w}, \cdots, v_{i-1}, v_{i+1}, \cdots, v_{i+w}\}$ is the set of neighbors of $v_i$ in a given random walk within window size $w$, the optimization problem then is 
$$\min_\Phi-\log \mathbb{P}(\left\{v_{i-w}, \cdots, v_{i-1}, v_{i+1}, \cdots, v_{i+w}\right\} | \phi\left(v_{i}\right))$$



\mpara{\app\cite{zhou2017scalable}.} 
As opposed to many embedding methods which preserve symmetric proximities, which can be insufficient for some applications, here an asymmetric proximity preserving (\app) method both for undirected and directed graphs is proposed. It uses an approximate version of Rooted PageRank wherein several paths are sampled from the starting vertex using a restart probability. Each vertex $v\in V$ is given two representations, source and target, denoted as ${\phi_s(v)}$ and $\phi_t(v)$ respectively. The two representations are learnt such that the likelihood of a training pair $(u,v)$ in their respective source and target roles, $\mathbb{P}(v|u)$ is maximized. The likelihood is modelled as softmax as follows and is optimized using negative sampling.
$$\mathbb{P}(v|u)= \frac{\exp ({\phi_s(u)} \cdot {\phi_t(v)})}{\sum_{n \in V} \exp (\phi_s(u) \cdot \phi_t(n)} $$


\mpara{\nerd\cite{NERD}.} 
Node Embeddings Respecting Directionality (\nerd) exploits the fact that in directed graphs, the neighborhood of a node differs based on its role as a source or a target (destination) node. To model node similarity while preserving its role semantics NERD proposes an alternating random walk strategy similar to SALSA\cite{Lempel00thestochastic} which alternates between source nodes (hubs) and target nodes (authorities). Two embeddings per node are learnt such that the probability of observing the sampled neighbors (in their respective roles) from an alternating random walk is maximized. It differs from \app in the kind of the random walk employed to collect training pairs and is more suitable for graphs with prominent hub authority structure.



\mpara{\hope\cite{ou2016asymmetric}.} 
High-Order Proximity preserved Embeddings or \hope for short, is an embedding framework also designed for directed graphs and based on finding source and target node embeddings  while optimizing for various high-order proximity measures exist like Katz Proximity, Personalized Pagerank, Common Neighbour measure, Adamic-Adar, etc.  In particular it finds a low rank decomposition of a proximity measure (in this work we use Katz proximity) where the two factors correspond to source and target embeddings of a node. If $S$ represents the Katz proximity matrix, \hope learns source and target representation matrices $\mathbf{U^s}$ and $\mathbf{U^t}$ while optimizing the following objective 
\[ \min_{\mathbf{U^s},\mathbf{U^t} } {||\mathbf{S} - \mathbf{U^s} \cdot \mathbf{(U^t)}^\top ||^2} .\]

\mpara{Unsupervised \graphsage\cite{hamilton2017inductive}.} \graphsage belongs to the family of Graph convolution based models \cite{kipf2017semi} which incorporate neighborhood aggregation mechanism in the learning algorithm to generate node representations. In this work we  use \graphsage in an unsupervised setting, using a graph-based loss which encourages nearby nodes to have similar representations, and far away nodes are enforced a distinct representation. In particular, it samples node pairs as in \deepwalk and uses the negative sampling based loss to embed vertices occuring together in short random walks closer.

\mpara{Graph Autoencoders.} This family of models aim to map each node to a vector in a lower dimension, from which reconstruction of the adjacency list of node should be possible. While Graph autoencoders using graph convolution networks(GCNs)\cite{kipf2017semi} as encoders are quite popular, recent empirical evidence \cite{salha2019keep} suggests that the GCN based encoders can be replaced by linear models without compromising performance in various downstream tasks. Though we compare various graph autoencoder models, namely GCN Autoencoder\cite{kipf2016variational}, GCN Variational Autoencoder\cite{kipf2016variational}, Linear Autoencoder, Linear Variational Autoencoder, and the deep versions of these autoencoders, we present results only corresponding to Graph Linear Autoencoder (\glae) as it showed much superior performance.



\subsection{ Rationale behind chosen Methods}
\label{sec:rational}
\mpara{Node Classification and State value functions.}
We choose \deepwalk in our study because of its robust performance in generating unsupervised node representations suitable for node classification task as shown in \cite{khosla2019comparative}. Note that as the task of finding an optimal policy is equivalent to finding optimal value function over nodes, one can interpret value functions as continuous forms of classes where similar nodes would have similar value functions. We hypothesize that methods performing well on node classification tasks would also generate embeddings suitable for approximating the optimal policies well. Our hypothesis is validated by our empirical results in which \deepwalk turns out to be the best performing method for all studied undirected MDPs.

\mpara{Asymmetric Local Neighborhoods.} A crucial aspect for node representations in MDPs which is ignored by several representation learning methods could be the existence of asymmetrical local neighborhoods even for the cases when the underlying mazes are undirected. \app learns two embeddings per vertex in its source and destination roles respectively, stressing the fact that a node $x$ is similar to node $y$ in its destination role if $y$ is reachable from $x$ in a small number of steps. As placement of obstacles can induce asymmetrical local structures in MDPs, we chose APP as one of the compared models. Moreover, APP is a general method also applicable for directed graphs.

\mpara{Homophily and Graph Convolution Based Methods.} Recent theoretical works\cite{li2018deeper,nt2019revisiting} imply that graph convolution operation is a special form of Laplacian smoothing which mixes the features of a vertex and its nearby neighbors. The smoothing operation makes the  features  of  vertices  in  the  same  cluster similar,  thus greatly easing the classification task, which is the key reason why GCNs work so well in node classification tasks. 
Moreover the empirical evidence \cite{khosla2019comparative} further supports the fact the GCN based models best perform for high homophily networks where labels do not vary a lot over adjacent vertices. We argue that the real world decision processes do not always give rise to high homophilic graphs, in the sense two adjacent nodes might not always have similar value functions. In this work we experiment with unsupervised \graphsage and several GCN based autoencoders. 

\mpara{Encoding Directionality.} For directed MDPs, it is crucial to take into account of the edge directions to decide on a favourable policy. In this work we therefore compare three methods specifically designed to learn unsupervised node representations for directed graphs. We chose \app, \nerd and \hope which exploit different characteristics  of directed graphs. Moreover, we also compare these methods with \deepwalk (used while ignoring the edge direction) to verify the importance of taking into account of the edge directionality while learning input state representations.

\section{Experiments and Results}
 
 \subsection{Experimental Setup}
 
\mpara{Grid-world Domains.} The experiments were carried out on various grid-world domains, more specifically $20\times 20$ or a 400 state square grids. These tasks are representative of a large class of discrete RL environments, and thus provide a suitable testbed. 
The domain has four walls, some obstacles, a start-state and a reward-state. The goal of the agent is to navigate the grid and reach the reward state. At each step, the agent gets some reward between $-1$ and $1$. Moreover, the obstacles in the domain are soft, meaning the agent can get into an obstacle state, but it incurs a heavy negative reward. The agent gets a reward of $+1$ if it reaches the reward state, $-0.3$ if it goes into an obstacle, and $-0.1$ if it bumps into a wall along the edge of the maze. Additionally to motivate the agent to find the shortest path, every valid move of the agent has a small negative reward of $-0.01$. We compare all GRL methods with a \textbf{matrix representation of the grid-world} where a grid is  a $20 \times 20$ matrix with all valid states represented by $1$, all obstacles by $0$, the target by $0.75$ and current agent-location by $0.75$.

Additionally, we tested our agent on two different types of grid-worlds. One, in which it resembles an undirected graph, meaning the agent can go into any of the empty neighbouring cells. The other type are the ones in which the agent's movements are restricted, adding directionality to the MDP graph. We describe different mazes as well as the corresponding results in Section~\ref{sec:results}.

\mpara{Model architecture employed for DQN.} The neural network has an input layer, which takes in the embedding $\phi(s)$ of dimension $d$ of the current state of the agent. For the matrix representation, the input layer is a flattened matrix corresponding to the grid. Further, there are two hidden layers with tanh activation, and then an output layer (with tanh activation) with four neurons. Each neuron corresponds to one of the four directions that the agent can move, and its value represents the state-action value of taking the respective action. At each time-step, the agent chooses the direction having the maximum value as predicted by the network. The loss function for the network is Mean Squared Error (MSE) loss.

\mpara{Hyperparameter Settings for DQN.}
During the training of the network, the agent performs episodes from the starting state till either it reaches the end state, or reaches a threshold of negative reward of $-25$. This high negative threshold ensures that the agent explores the maze long enough to learn. After completing an episode, the agent is reset to the initial state. The agent learns using experience replay. So during the agent's episodes, it's steps are stored in a memory in the form of $(s_t,a,r_t,s_{t+1})$ (where the symbols have usual meaning). Moreover, the 'experience-gaining' phase is separate from the 'learning phase', even though they are interleaved. After each time step, random previous actions from the memory are selected and are used in the learning phase. 

\mpara{Hyperparameter Settings for GRL Methods.} 
The representation learning phase requires the graph as an input in either edgelist or adjacency list format. For \deepwalk the walk-length, and window-size were set to $20$ and $5$ respectively. For \graphsage, the GCN mean aggregator was used. For all other methods like \app, \nerd and \glae, the default hyperparameters for these algorithms were used. In methods like \app, \nerd and \hope for example, which generate two embeddings, each of dimension $d$ per node of a graph, the embeddings of current state and the next state (corresponding to a particular action) are concatenated to form an input embedding of dimension $2d$. The rest of the architecture of the network remains the same.

\mpara{Evaluation Metrics}
We used standard metrics for evaluation which are, average cumulative reward received by the agent against number of time-steps, and against the number of episodes completed by the agent. As discussed before, the goal of the agent is to accumulate the maximum reward from the grid. The better the method, the quicker the agent reaches the target and the more reward it accumulates. We argue that a better method would represent the input state-space better, and hence the agent would accumulate more reward. Another way to look at this is that a better input state representations would result in less number of episodes, or less number of steps to find out the optimal path to reach the target. With this in mind, the different domains which were used, and the plots of the metrics discussed above are presented in the next section.

\subsection{Results on Sampled Complete Grid Environments}
\label{sec:results}
In this section we describe the results for different types of mazes. Here we assume that enough number of samples were drawn in order to build the complete mazes. In addition the empirical time complexity of maze construction from the acquired samples is roughly $1\%$ of the time for training the deep network. We observe that for all the studied mazes the pre-trained state representations outperform the matrix based state representation. 
 We note that overall we compared $9$ GRL methods out of which results for $6$ are presented in this section. The other three methods, which are GCN based shallow and deep autoencoders showed worse results than the others and hence we don't show their performance here.

As already mentioned we plot the standard evaluation metrics: (i) average cumulative reward against the number of time-steps (ii) average cumulative reward against the number of episodes with $90\%$ confidence intervals. For the results presented in this section we run all methods for  60 episodes and repeat experiments for atleast 40 iterations. We note that in the plots corresponding to time steps, some of the methods stop much earlier than the others, which indicates that they have shorter episodes, which in turn indicates that they were much faster in finding the optimal policy and achieving higher rewards much earlier.




\subsubsection{Maze 1.} 
\begin{figure}[h!]
    \centering
     \includegraphics[width=0.5\textwidth]{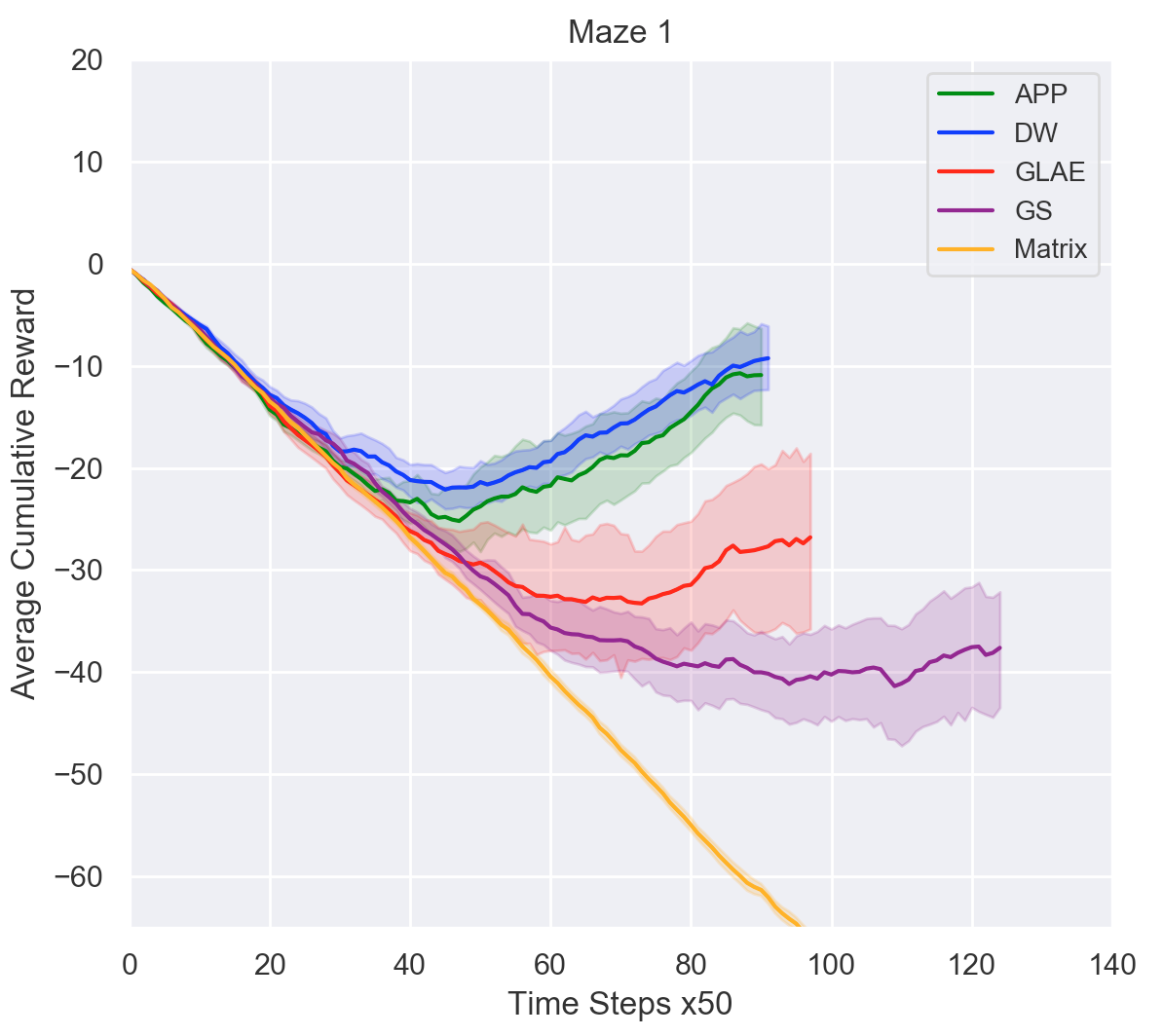} \hfill
    \includegraphics[width=0.45\textwidth]{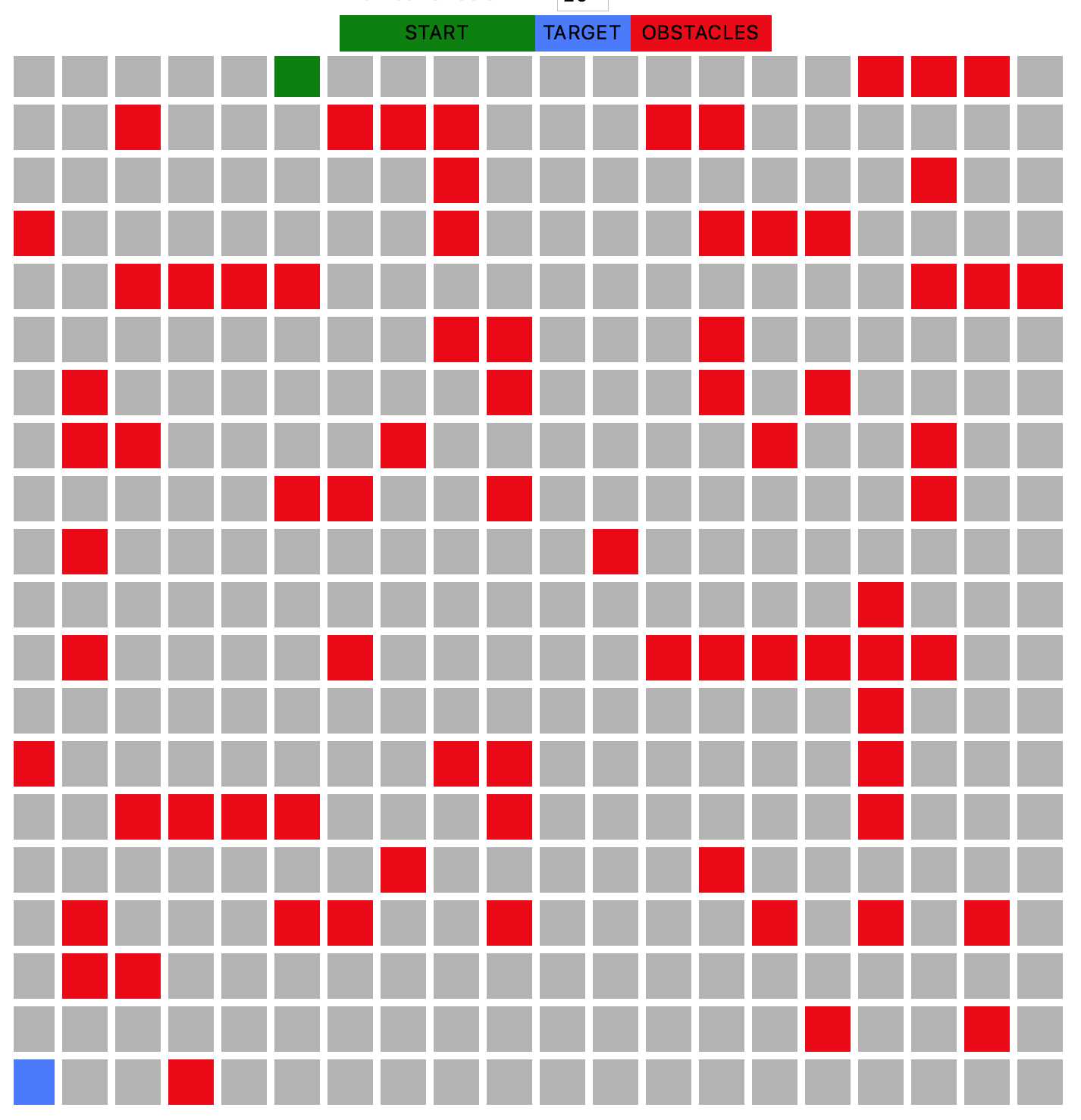}
    \caption{Maze 1 against its plot of the average episode-wise reward obtained with a 90\% confidence interval.}
    \label{fig:Maze1}
\end{figure}

\begin{figure}[h!]
    \centering
     \includegraphics[width=0.55\textwidth]{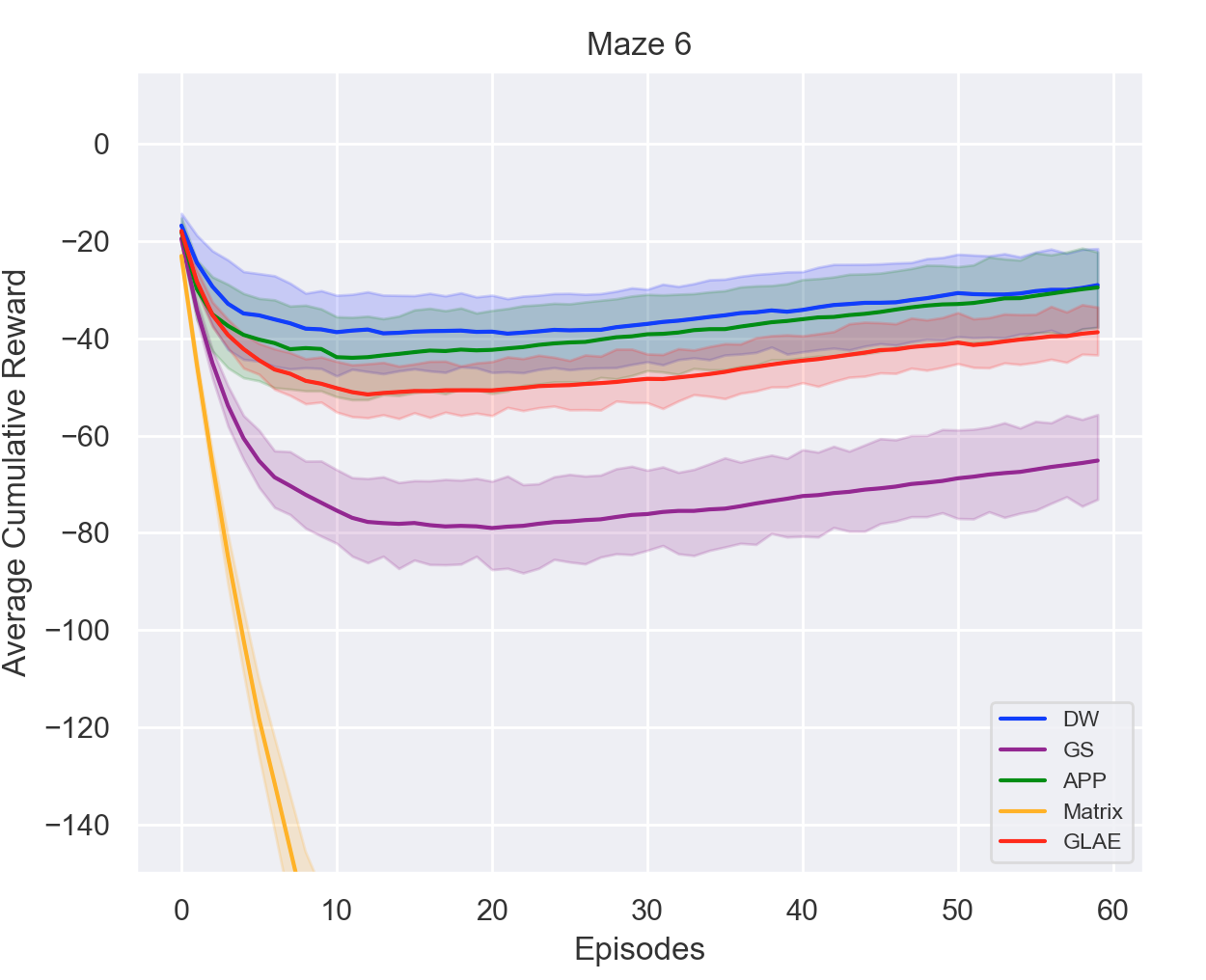}
    \caption{Episode-wise rewards obtained for maze 1}
    \label{fig:Maze1EW}
\end{figure}

The obstacle and the start state are in farthest rows of the maze.The results are presented in Figures \ref{fig:Maze1} and \ref{fig:Maze1EW}. \deepwalk and \app outperform other GRL methods.  While Figure~\ref{fig:Maze1} shows the cumulative reward against time-steps, Figure~\ref{fig:Maze1EW} shows it against number of episodes. 
For the rest of the mazes we will stick only to the step-wise plots as they clearly are more informative than the episode-wise plots. 

\subsubsection{Maze 2.} 
\begin{figure}[h!]
    \centering
     \includegraphics[width=0.50\textwidth]{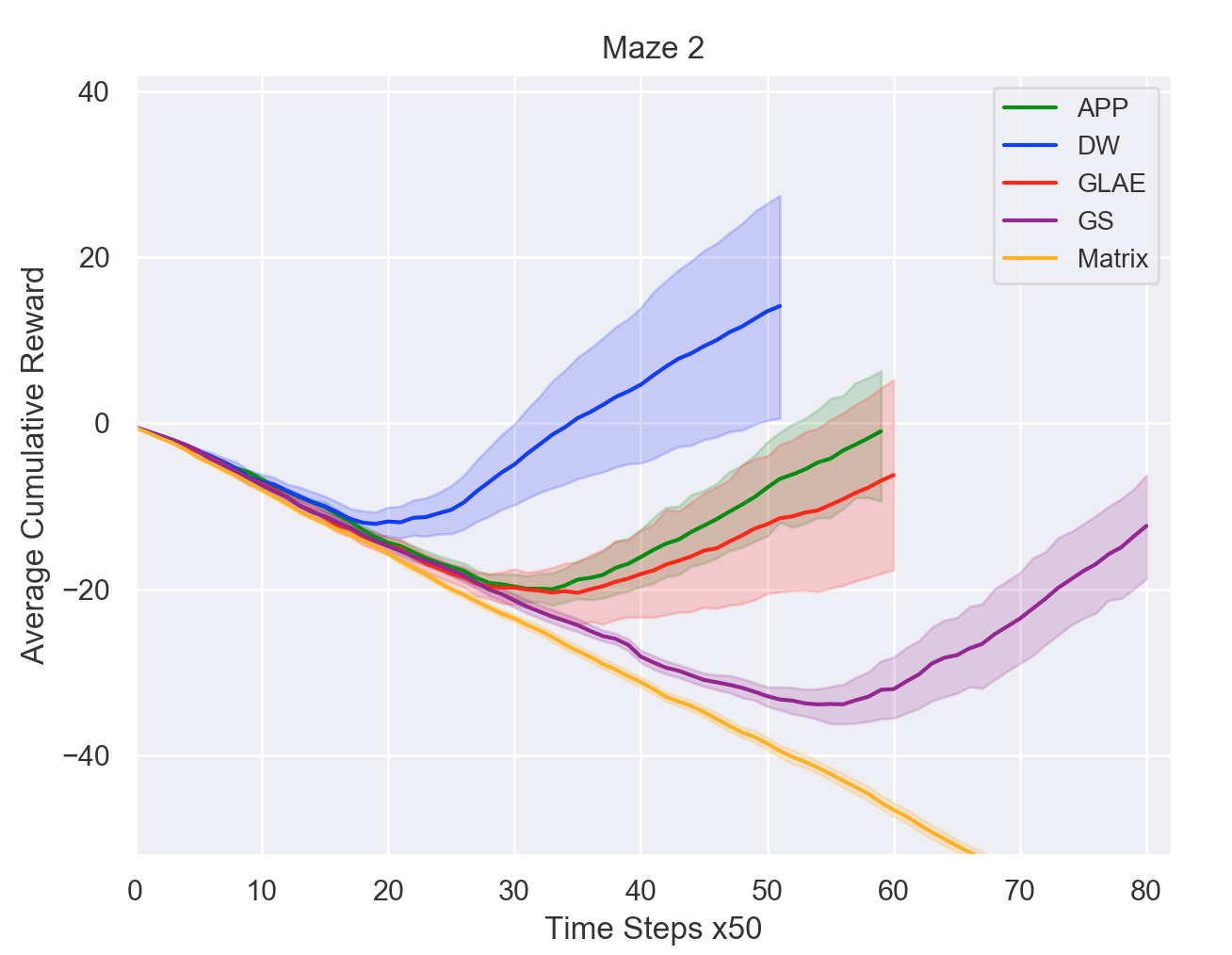} \hfill
    \includegraphics[width=0.40\textwidth]{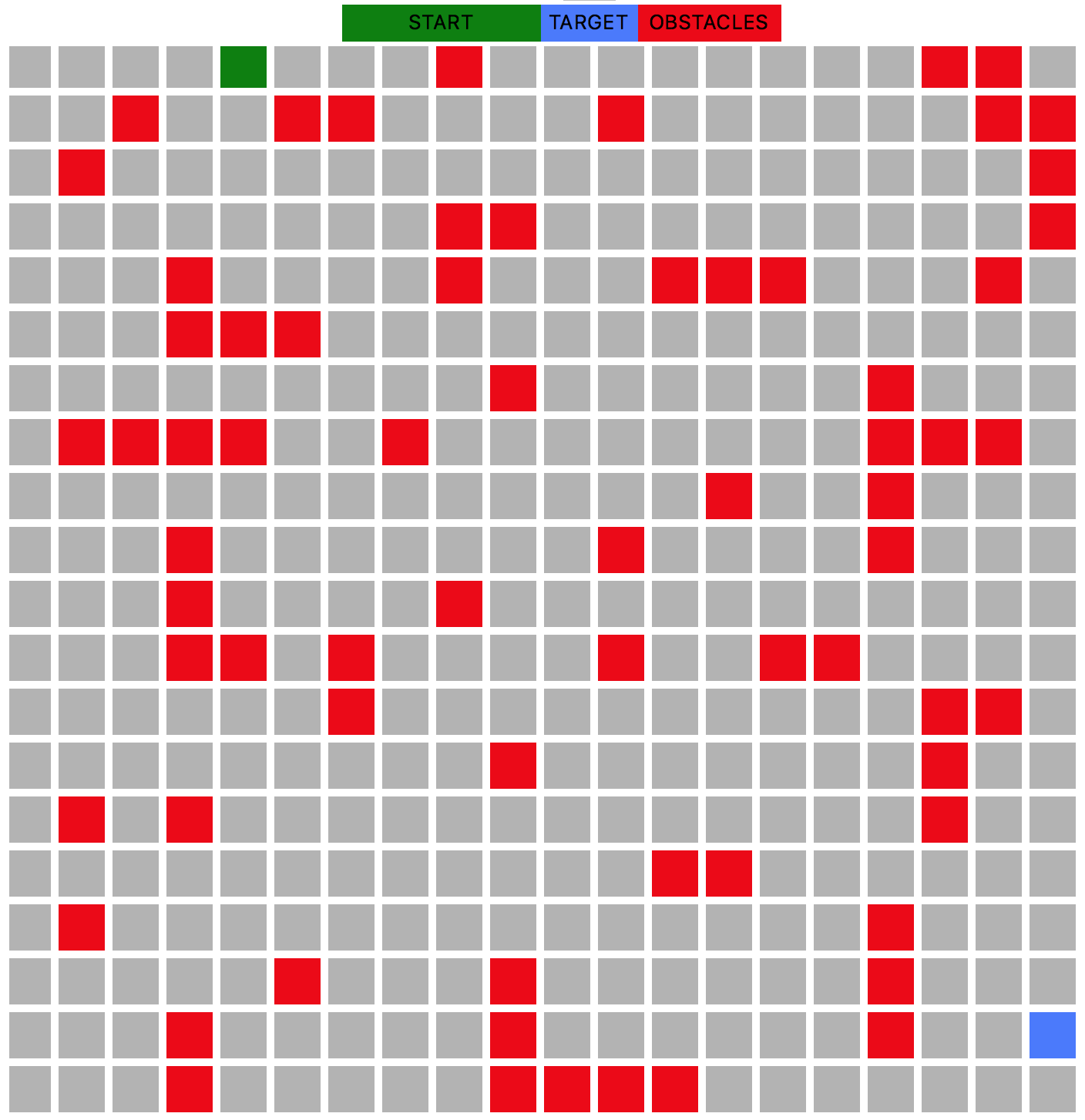}
    \caption{Maze 2 against its plot of the step-wise reward obtained.}
    \label{fig:Maze2}
\end{figure}

In this maze the agent is expected to find a longer but less convoluted path than in maze 1. The results for this domain are presented in Figure \ref{fig:Maze2}. Once again we observe that \deepwalk is the best performing method followed by \app and \glae. 

\subsubsection{Maze 3.} 
\begin{figure}[h!]
    \centering
     \includegraphics[width=0.50\textwidth]{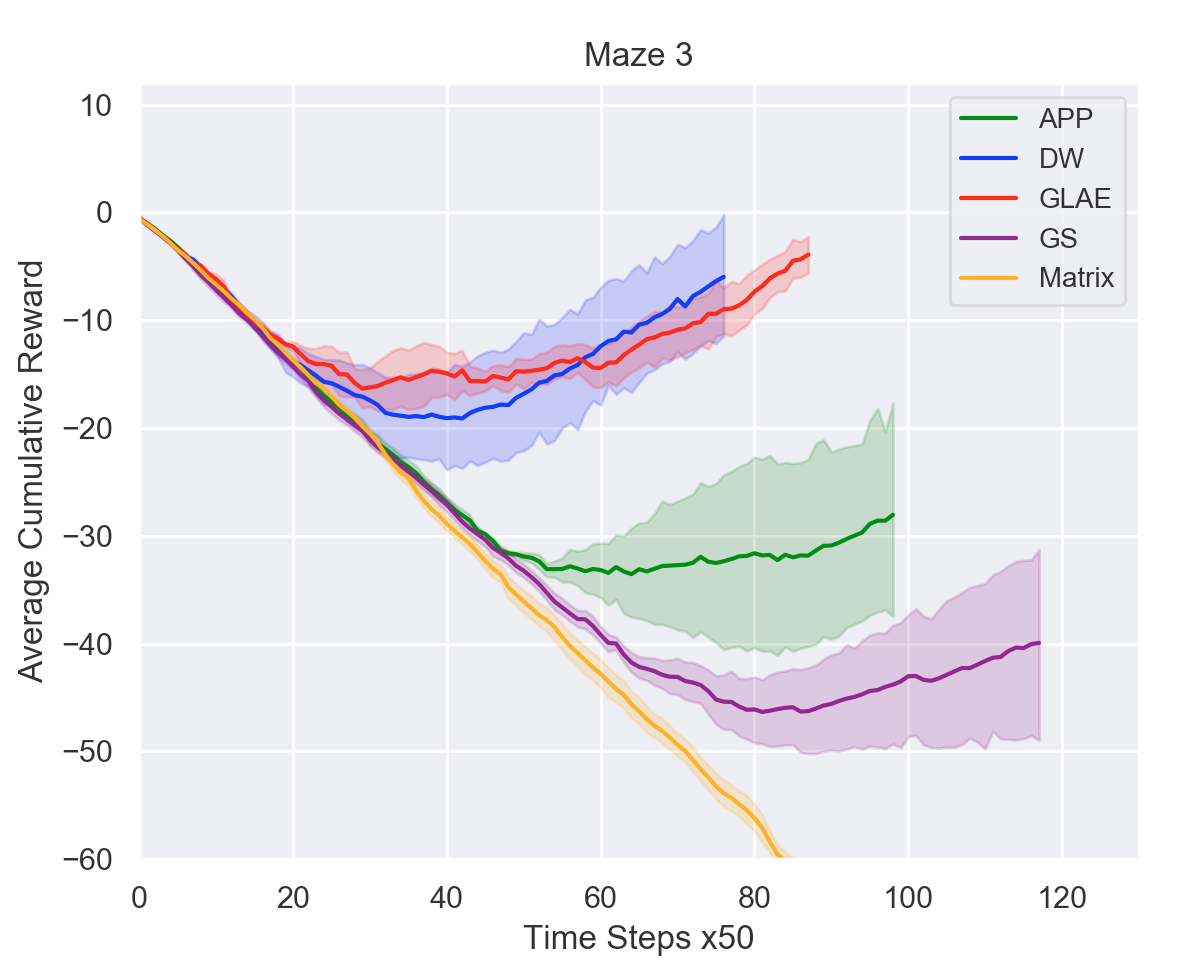} \hfill
    \includegraphics[width=0.40\textwidth]{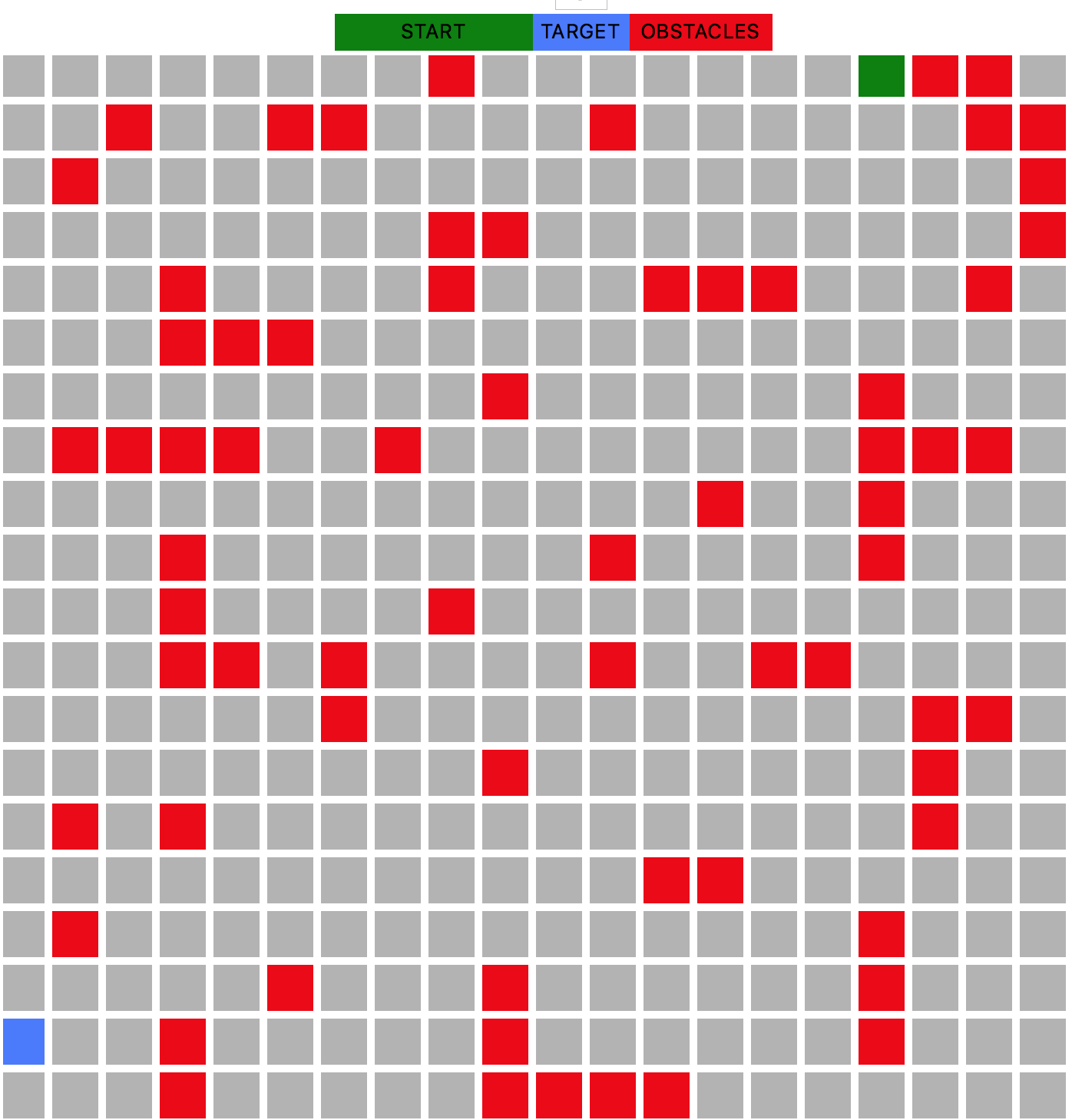}
    \caption{Maze 3 against its plot of the step-wise reward obtained.}
    \label{fig:Maze3}
\end{figure}

This domain is very similar to maze 2 in that the obstacles are exactly at the same place, but the locations of the start state and the target are changed. The results are presented in Figure \ref{fig:Maze3}. As opposed to Maze 1, 
\glae outperforms \app and performs comparable to \deepwalk.

\subsubsection{Mazes 4 and 5.}
Mazes 4 and 5 are directed in nature. Maze 4 is like Maze 3 in terms of obstacles location, but only down and left actions of agent are permitted. Maze 5 is similar to Maze 1, but with $15\%$ of the possible actions removed randomly. For both these mazes, we used methods like \hope and \nerd, which are explicitly designed for directed graphs. We additionally compared them with \deepwalk (while ignoring the edge directionality) to investigate if the edge directionality has a large impact on state representations. 
In Maze 4, \app and \deepwalk are the best performing methods followed by \nerd, whereas in Maze 5, \app and \nerd outperform \deepwalk.


\begin{figure}[h!]
    \centering
     \includegraphics[width=0.45\textwidth]{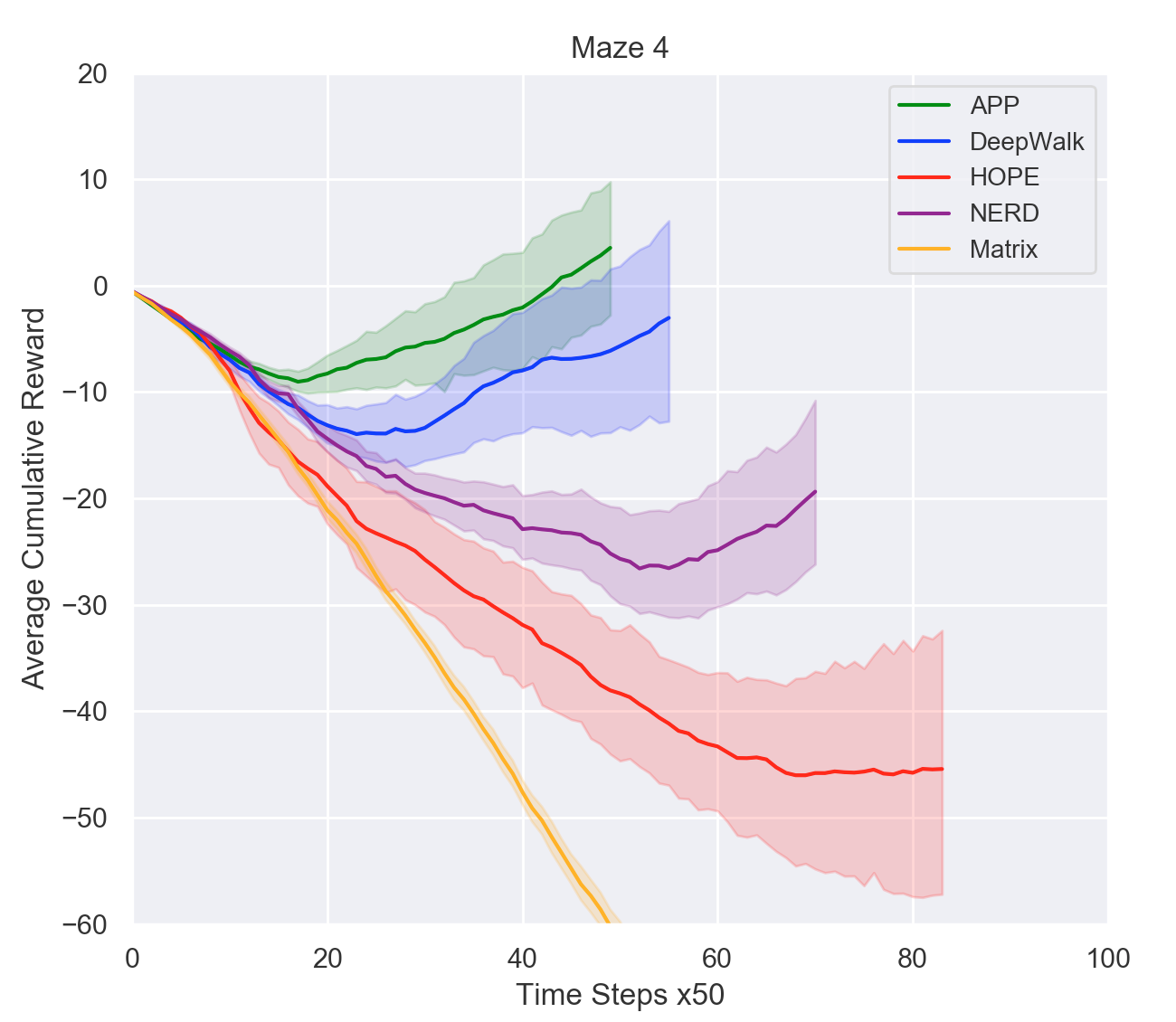} \hfill
      \includegraphics[width=0.43\textwidth]{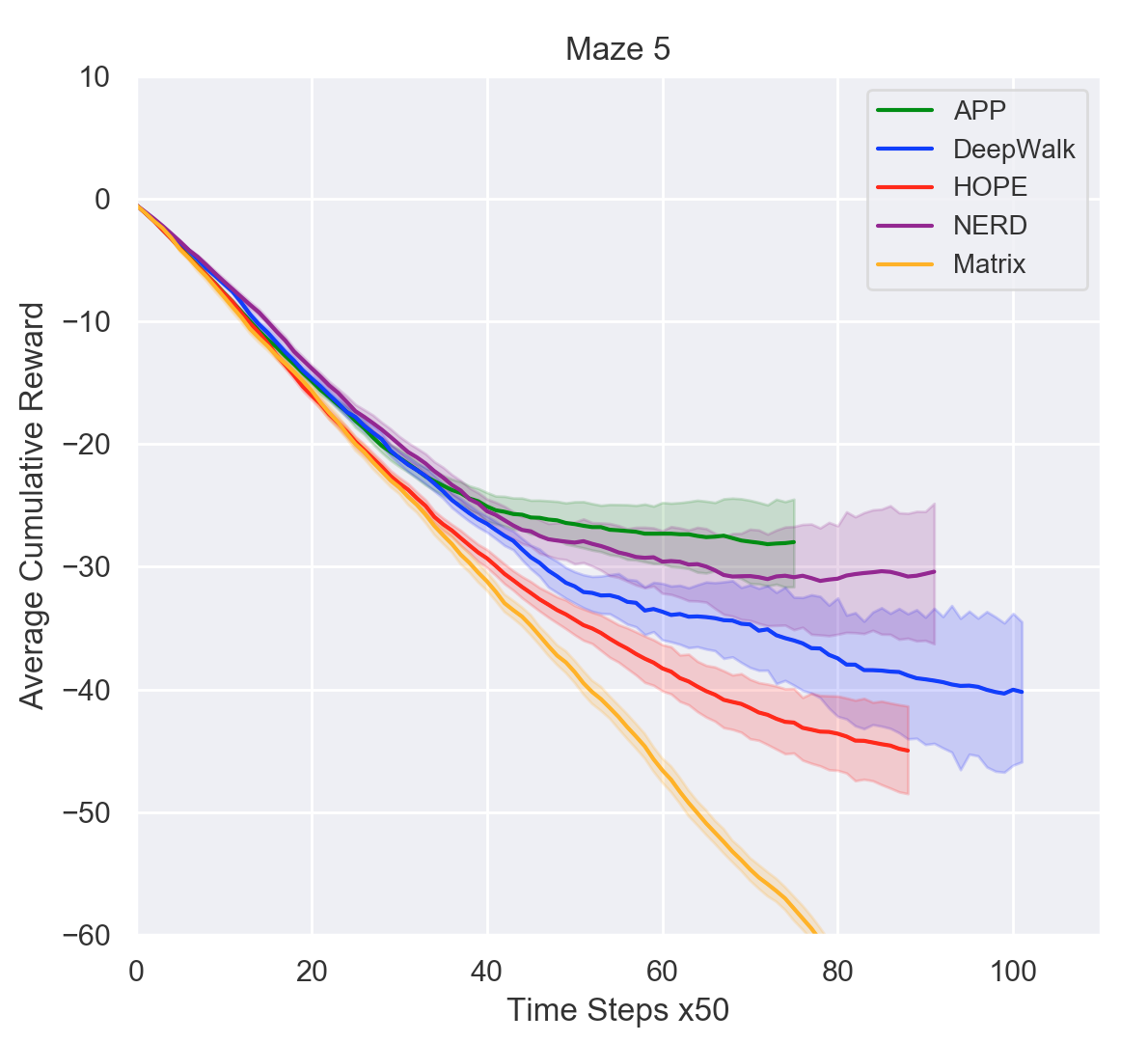}
    \caption{Step-wise reward obtained for maze 4 and maze 5}
    \label{fig:Maze45}
\end{figure}

\subsubsection{Main Observations.} With respect to our results in this section, we make the following main observations and conclusions.
\begin{enumerate}
    \item For undirected mazes \deepwalk is the best method. \app and \glae are also competitive methods but their performance varies over different mazes. All these methods outperform \graphsage and the raw matrix input representation. 
    \item The graph convolution based method,\graphsage is the worst performing method for all undirected mazes among the GRL methods. We remark that in our initial experiments we had also compared GCN based shallow and deep autoencoders and found that the graph linear autoencoder was the better performing model. We, therefore, conclude that GCN based methods are not the best choice for learning unsupervised state representations in MDPs in absence of external node features. 
    \item For directed mazes, \app is the best performing method followed either by \deepwalk or \nerd. The good performance of \deepwalk in these mazes is a bit surprising and points to the fact that the directionality of the underlying MDPs (at least for the studied environments) can be ignored as far as learning state representations is concerned.
    \item We also observe that the best performing methods also incur shorter episodes supporting the fact that a better input state representation not only improved the learning quality but also made it faster.
\end{enumerate}

\subsection{Sensitivity to Embedding Dimension}
\begin{figure}[h!]
\begin{subfigure}{.5\textwidth}
\centering
\includegraphics[width=\textwidth]{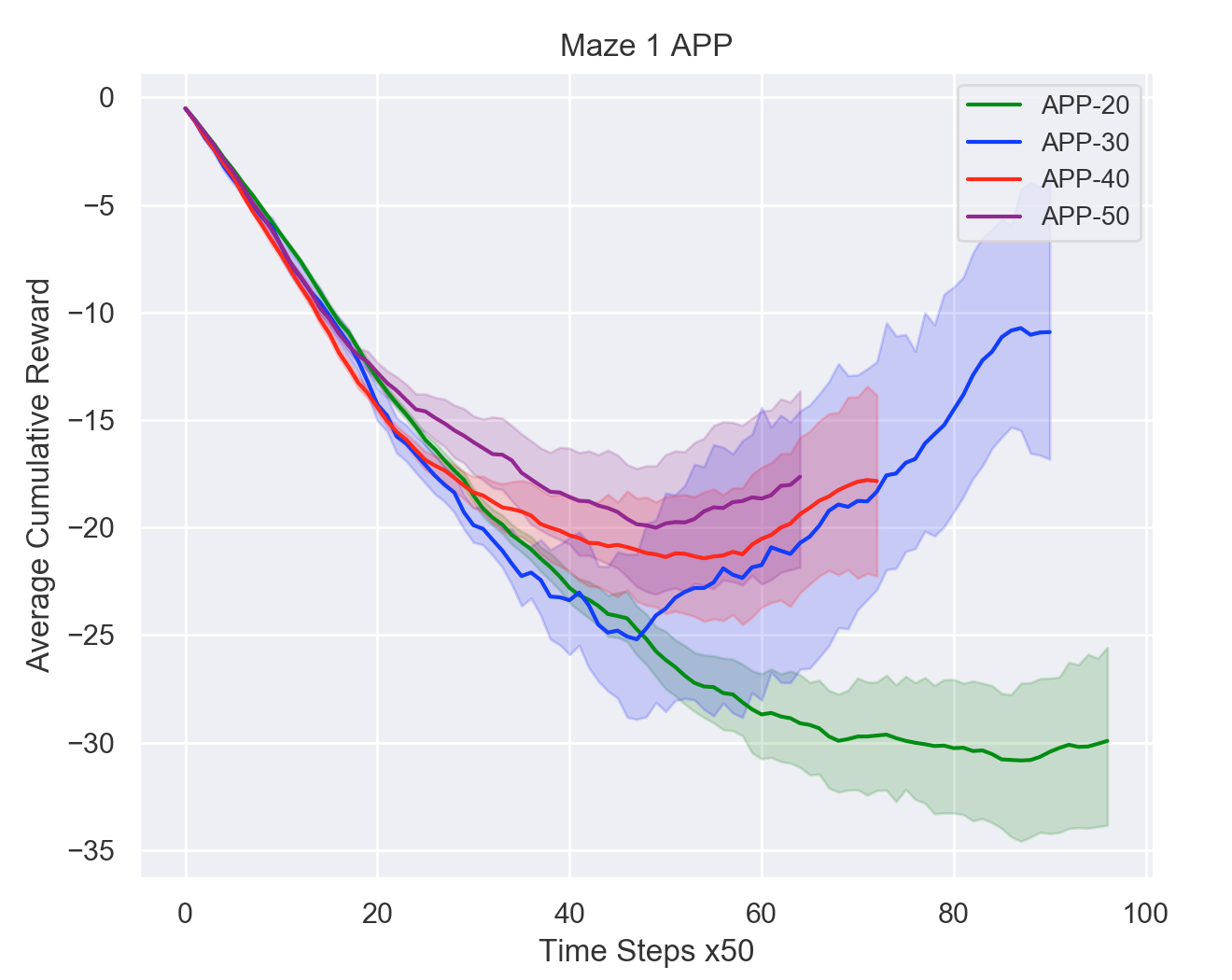}
\caption{\app}
\label{fig:Maze1DimAPP}
\end{subfigure}
\begin{subfigure}{.5\textwidth}
\centering
\includegraphics[width=\textwidth]{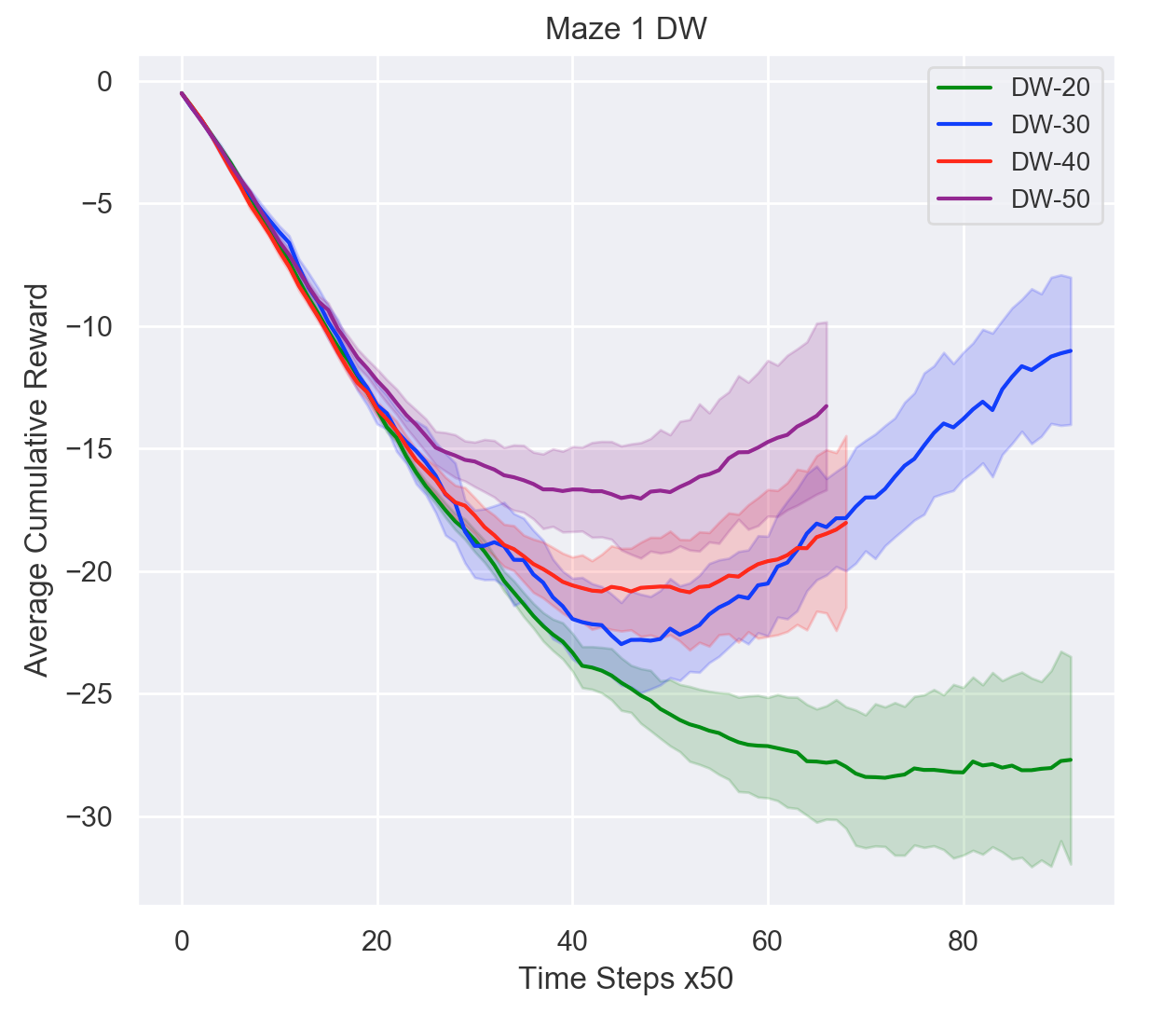} 
\caption{\deepwalk}
\label{fig:Maze1DimDW}
\end{subfigure}
\begin{subfigure}{.5\textwidth}
\centering
\includegraphics[width=0.95\textwidth]{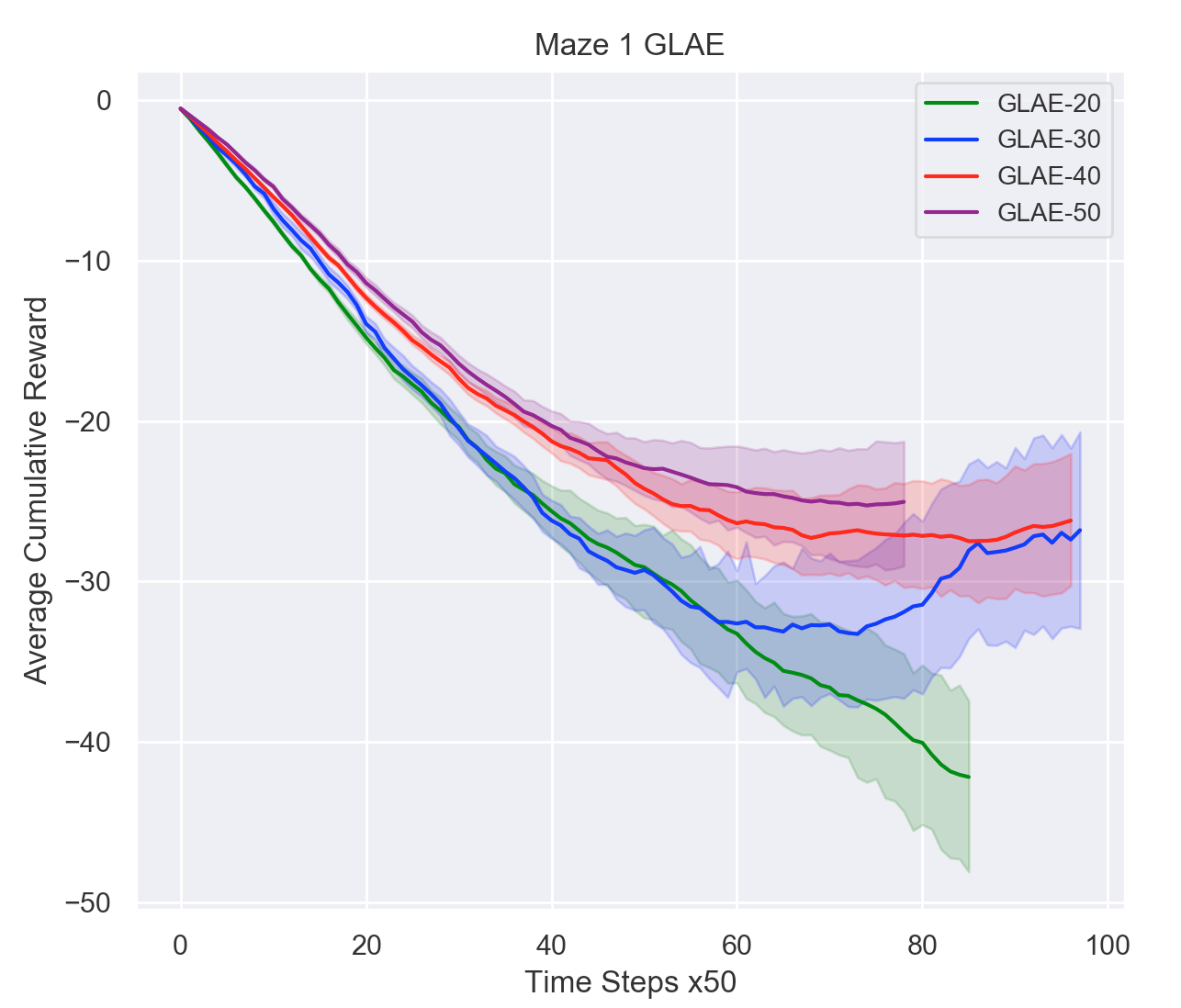} 
\caption{\glae}
\label{fig:Maze1DimGLAE}
\end{subfigure}
\begin{subfigure}{.5\textwidth}
\centering
\includegraphics[width=1\textwidth]{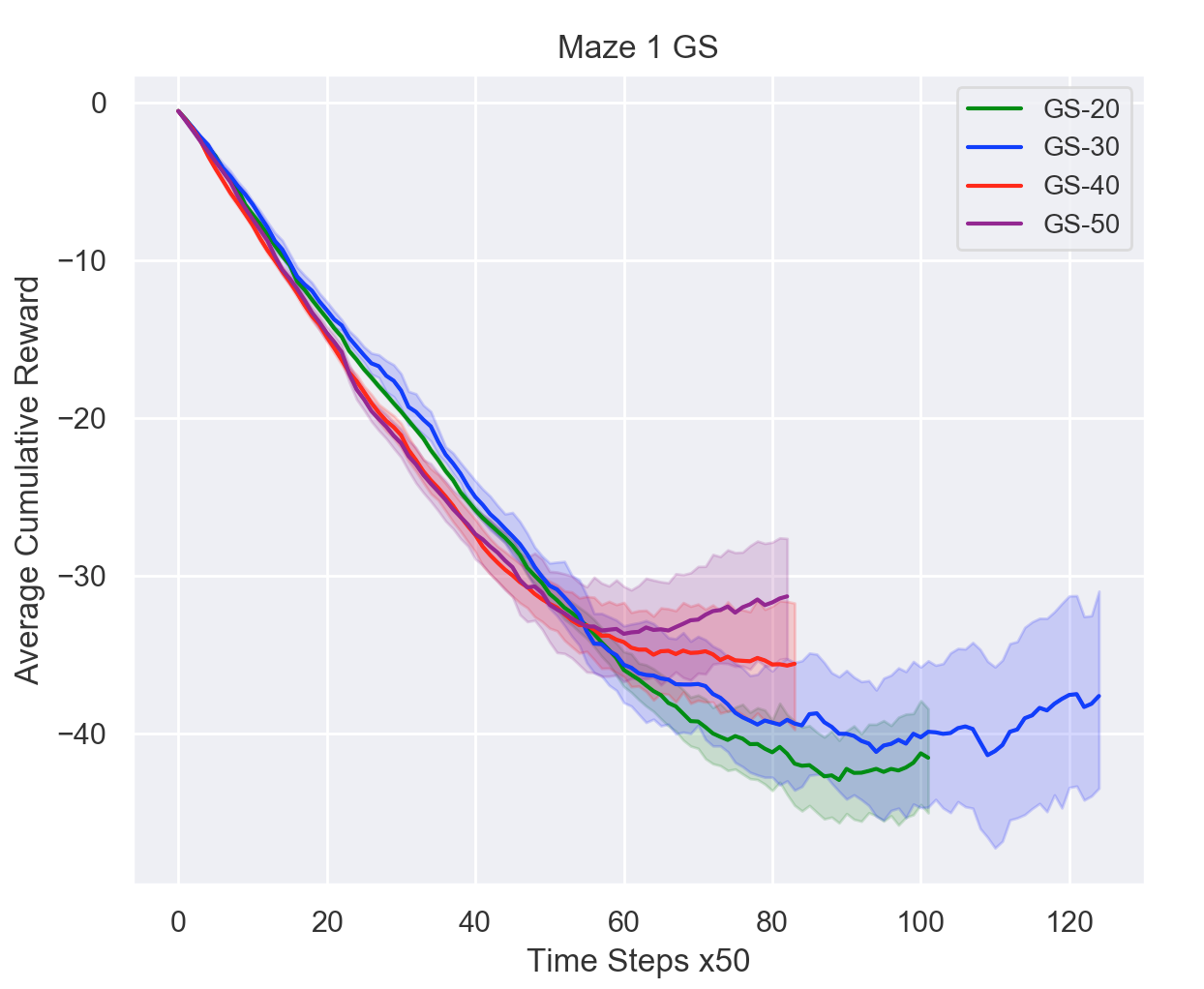}
\caption{\graphsage}
\label{fig:Maze1DimGS}
\end{subfigure}
\caption{Sensitivity to dimensions. All results correspond to experiments of Maze 1 using different embedding dimensions, $d\in\{ 20,30,40,50\}$.}
\label{Maze1Dim}
\end{figure}

In this section we study the effect of dimension of embeddings on learning performance of the DQN.
In principle, we want the dimension of the embedding space to be considerably smaller than the original dimension of the state-space. In Figure \ref{Maze1Dim} we consider four different dimensions for 4 different methods. For all these methods, we observe a diminishing return after a dimension of size 30. While increasing the dimension would up to some extent also improve performance, but it would incur additional time cost of training extra parameters. For \app and \deepwalk we observe that a higher number of dimensions only lead to smaller episodes. 


\subsection{Sensitivity to Sample Size}
\begin{figure}[h!]
\begin{subfigure}{.5\textwidth}
    \centering
    \includegraphics[width=0.95\textwidth]{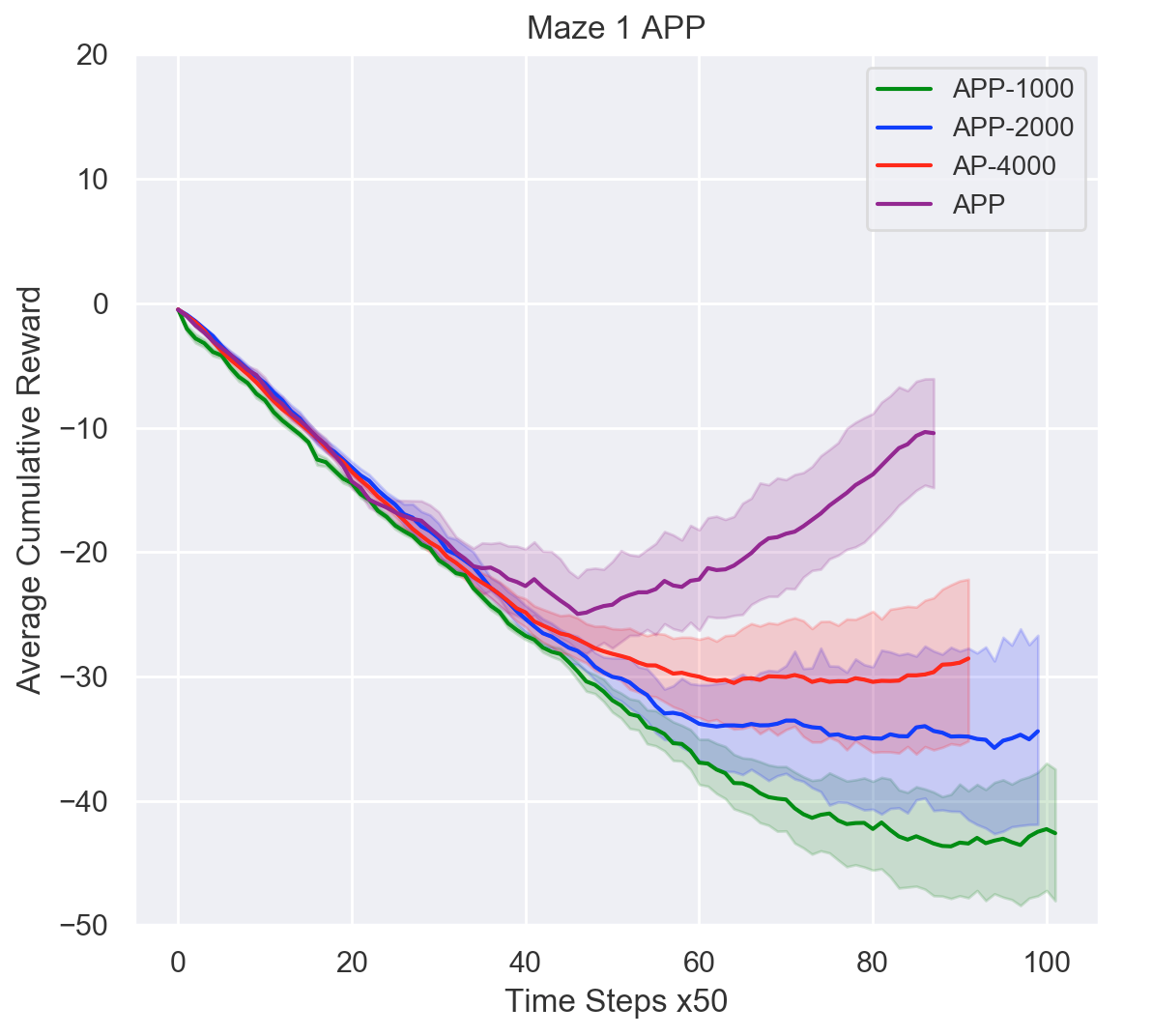}
    \caption{\app}
    \label{fig:Maze1DimAPP}
\end{subfigure}
\begin{subfigure}{.5\textwidth}
    \centering
    \includegraphics[width=1\textwidth]{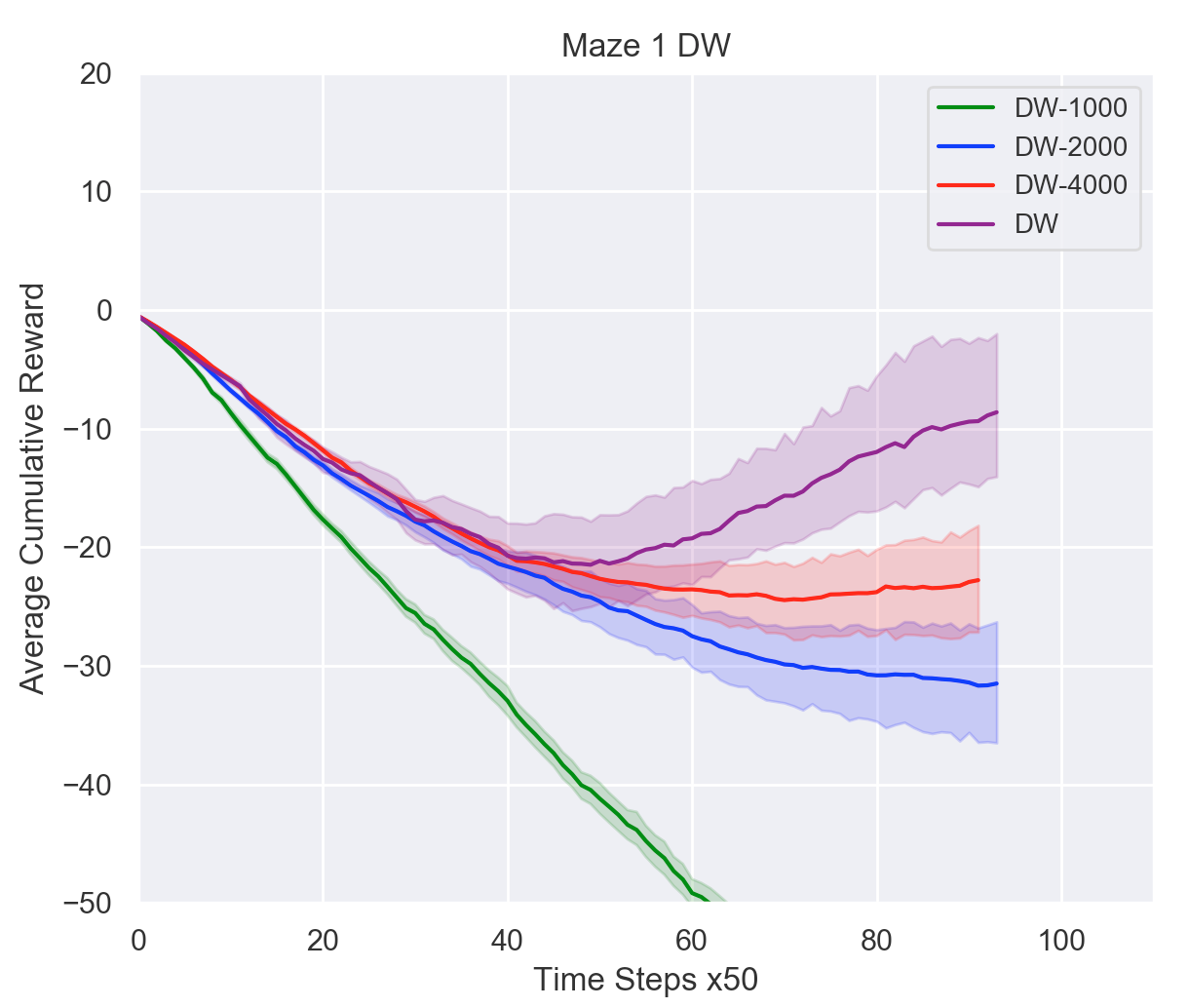} 
    \caption{\deepwalk}
    \label{fig:Maze1DimDW}
\end{subfigure}
\begin{subfigure}{.5\textwidth}
    \centering
    \includegraphics[width=0.95\textwidth]{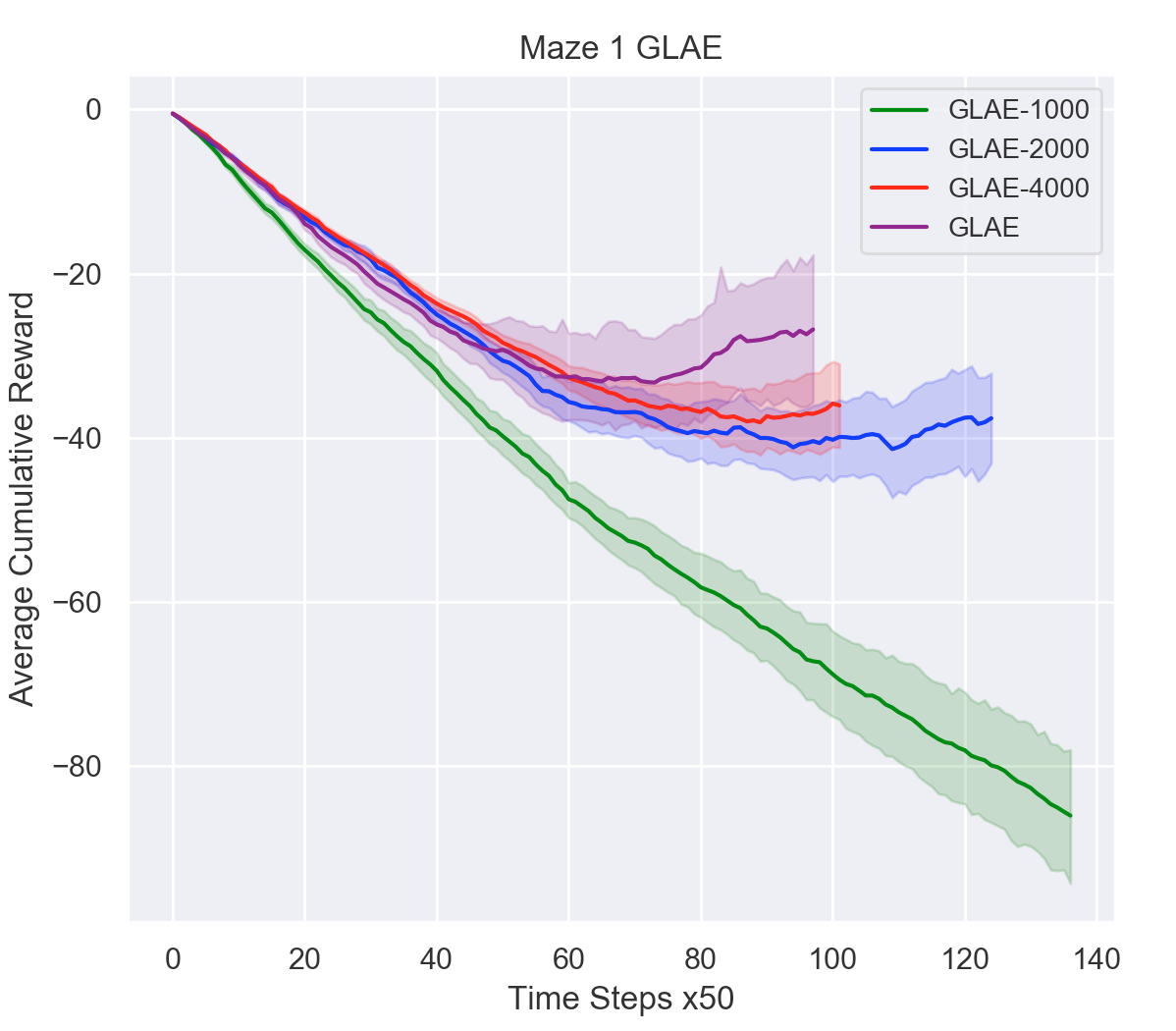} 
    \caption{\glae}
    \label{fig:Maze1DimGLAE}
\end{subfigure}
\begin{subfigure}{.5\textwidth}
    \centering
    \includegraphics[width=1\textwidth]{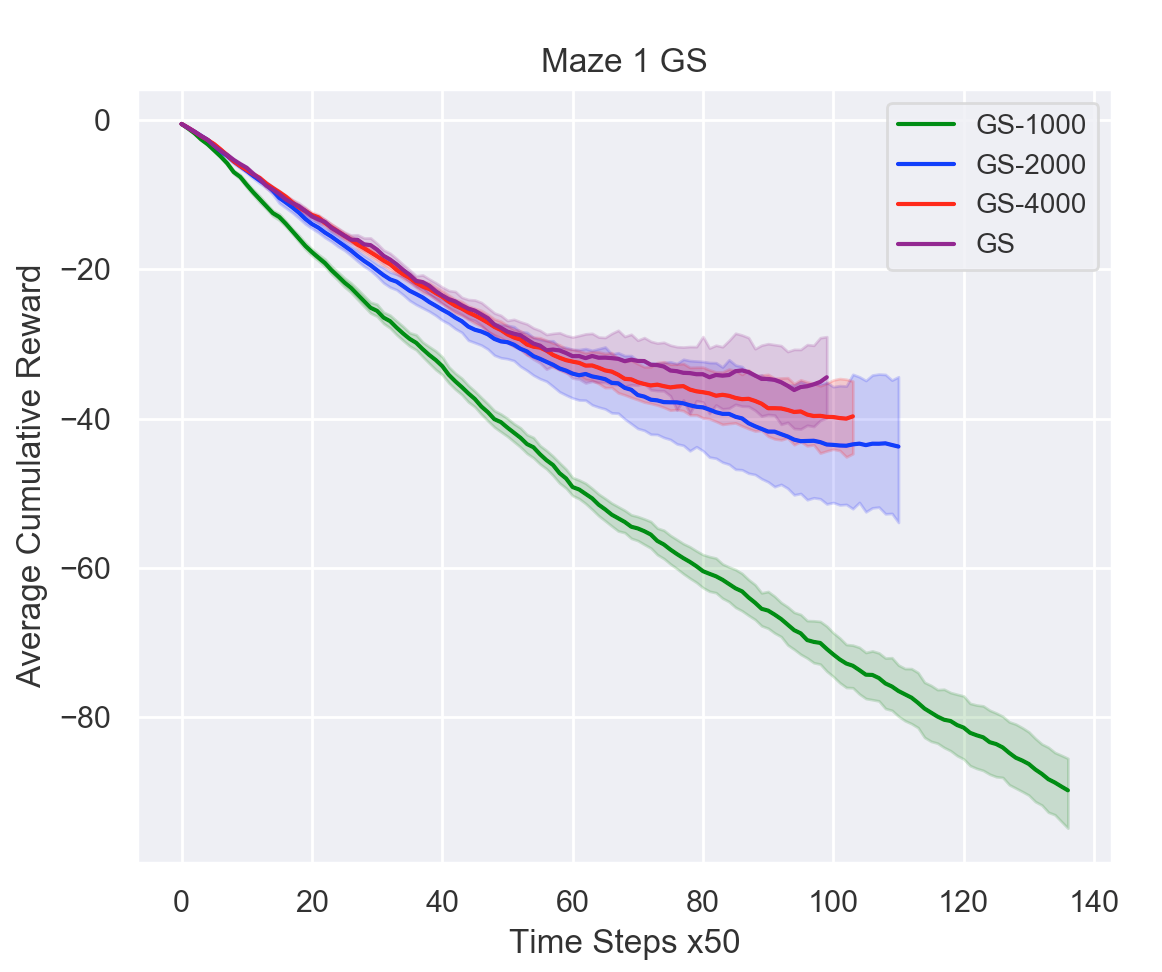}
    \caption{\graphsage}
    \label{fig:Maze1DimGS}
\end{subfigure}

\caption{Sensitivity to sample size. All results correspond to experiments of Maze 1 constructed using different sizes of random samples, $S\in\{1000,2000,4000\}$. The purple line corresponds to when complete maze was used.}
\label{Maze1Sample}
\end{figure}

As described previously, in all the results until now, we assume that enough samples were drawn to build the complete maze. In this section, we investigate how sample size affects the learning. On average, to reconstruct a $20 \times 20$ maze by drawing random samples, approximately $8000$ such samples are required. The results are shown in Figure~\ref{Maze1Sample}. The purple line shows the cumulative reward when the complete maze is used to learn state representations. From the plots, we see that the green line, corresponding to $1000$ samples performs the worst, and the cumulative rewards never start to rise. On increasing the sample sizes, the performance incrementally improves. For \graphsage and \glae in particular, even with a small sample size of $2000$ show performance comparable to collecting $8000$ samples, though their overall performance is worse than \app and \deepwalk.

\section{Conclusion}
In this work we propose and evaluate a wide range of graph based representation learning approaches to generate state features based on topological structure of MDPs, leading to improved learning performance in navigational tasks. In particular, we conducted an empirical study over a wide range of unsupervised graph representation learning methods and conclude that a random walk based method \deepwalk is best suited for generating low dimensional state representations based on the topological structure of the underlying MDP. For directed MDPS, \app and \nerd are competitive methods and can offer advantages in encoding specific directed MDPs. We also find that the more popular GCN based models are the worst performing among the compared models. We also performed parameter sensitivity experiments where we investigated the effect of (i) increasing embedding dimension (ii) increasing the sample sizes to generate the MDP. We observe that though increasing dimensions improve the learning performance, a small value of $30$ for embedding dimensions already suffices to obtain competitive performance. This is important for DQNs where the number of network parameters increase with the number of input embedding dimensions. We also show that the representation learning methods also perform comparably to their best performance (computed over the complete MDP) while using a much smaller number of random samples to build an estimate of the MDP. Our work also shows that RL can serve as a promising application and test bed for graph representation learning approaches.

\bibliographystyle{splncs04}
\bibliography{references}
%




\end{document}